\documentstyle[aclap]{article}

\newtheorem{example}{Example}

\title{\vspace{-0.5in}Computing Optimal Descriptions for Optimality Theory Grammars
with Context-Free Position Structures}
\author{Bruce Tesar \\
The Rutgers Center for Cognitive Science /\\
The Linguistics Department\\
Rutgers University\\
Piscataway, NJ 08855 USA \\
{\tt tesar@ruccs.rutgers.edu}}

\begin{document}
\maketitle

\begin{abstract}
This paper describes an algorithm for computing optimal structural
descriptions for Optimality Theory grammars with context-free position
structures. This algorithm extends Tesar's dynamic programming approach \cite
{Tesar:ParsingTR94} \cite{Tesar:ParsingTR95} to computing optimal structural
descriptions from regular to context-free structures. The generalization to
context-free structures creates several complications, all of which are
overcome without compromising the core dynamic programming approach. The
resulting algorithm has a time complexity cubic in the length of the input,
and is applicable to grammars with universal constraints that exhibit
context-free locality.
\end{abstract}

\section{Computing Optimal Descriptions in Optimality Theory}

In Optimality Theory \cite{Prince+Smolensky:OT}, grammaticality is defined
in terms of optimization. For any given linguistic input, the grammatical
structural description of that input is the description, selected from a set
of candidate descriptions for that input, that best satisfies a ranked set
of universal constraints. The universal constraints often conflict:
satisfying one constraint may only be possible at the expense of violating
another one. These conflicts are resolved by ranking the universal
constraints in a strict dominance hierarchy: one violation of a given
constraint is strictly worse than any number of violations of a lower-ranked
constraint. When comparing two descriptions, the one which better satisfies
the ranked constraints has higher Harmony. Cross-linguistic variation is
accounted for by differences in the ranking of the same constraints.

The term {\em linguistic input} should here be understood as something like
an underlying form. In phonology, an input might be a string of segmental
material; in syntax, it might be a verb's argument structure, along with the
arguments. For expositional purposes, this paper will assume linguistic
inputs to be ordered strings of segments. A candidate structural description
for an input is a full linguistic description containing that input, and
indicating what the (pronounced) surface realization is. An important
property of Optimality Theory (OT) grammars is that they do not accept or
reject inputs; every possible input is assigned a description by the grammar.

The formal definition of Optimality Theory posits a function, {\it Gen},
which maps an input to a large (often infinite) set of candidate structural
descriptions, all of which are evaluated in parallel by the universal
constraints. An OT grammar does not itself specify an algorithm, it simply
assigns a grammatical structural description to each input. However, one can
ask the computational question of whether efficient algorithms exist to
compute the description assigned to a linguistic input by a grammar.

The most apparent computational challenge is posed by the allowance of {\em %
faithfulness} violations: the surface form of a structural description may
not be identical with the input. Structural positions not filled with input
segments constitute overparsing (epenthesis). Input segments not parsed into
structural positions do not appear in the surface pronunciation, and
constitute underparsing (deletion). To the extent that underparsing and
overparsing are avoided, the description is said to be faithful to the
input. Crucial to Optimality Theory are faithfulness constraints, which are
violated by underparsing and overparsing. The faithfulness constraints
ensure that a grammar will only tolerate deviations of the surface form from
the input form which are necessary to satisfy structural constraints
dominating the faithfulness constraints.

Computing an optimal description means considering a space of candidate
descriptions that include structures with a variety of faithfulness
violations, and evaluating those candidates with respect to a ranking in
which structural and faithfulness constraints may be interleaved. This is
parsing in the generic sense: a structural description is being assigned to
an input. It is, however, distinct from what is traditionally thought of as
parsing in computational linguistics. Traditional parsing attempts to
construct a grammatical description with a surface form matching the given
input string exactly; if a description cannot be fit exactly, the input
string is rejected as ungrammatical. Traditional parsing can be thought of
as enforcing faithfulness absolutely, with no faithfulness violations are
allowed. Partly for this reason, traditional parsing is usually understood
as mapping a surface form to a description. In the computation of optimal
descriptions considered here, a candidate that is fully faithful to the
input may be tossed aside by the grammar in favor of a less faithful
description better satisfying other (dominant) constraints. Computing an
optimal description in Optimality Theory is more naturally thought of as
mapping an underlying form to a description, perhaps as part of the process
of language production.

Tesar \cite{Tesar:ParsingTR94} \cite{Tesar:ParsingTR95} has developed
algorithms for computing optimal descriptions, based upon dynamic
programming. The details laid out in \cite{Tesar:ParsingTR95} focused on the
case where the set of structures underlying the {\it Gen} function are
formally regular. In this paper, Tesar's basic approach is adopted, and
extended to grammars with a {\it Gen} function employing fully context-free
structures. Using such context-free structures introduces some complications
not apparent with the regular case. This paper demonstrates that the
complications can be dealt with, and that the dynamic programming case may
be fully extended to grammars with context-free structures.

\section{Context-Free Position Structure Grammars}

Tesar \cite{Tesar:ParsingTR95} formalizes {\it Gen} as a set of matchings
between an ordered string of input segments and the terminals of each of a
set of position structures. The set of possible position structures is
defined by a formal grammar, the {\em position structure grammar}. A
position structure has as terminals structural positions. In a valid
structural description, each structural position may be filled with at most
one input segment, and each input segment may be parsed into at most one
position. The linear order of the input must be preserved in all candidate
structural descriptions.

This paper considers Optimality Theory grammars where the position structure
grammar is context-free; that is, the space of position structures can be
described by a formal context-free grammar. As an illustration, consider the
grammar in Examples \ref{Ex:PSG} and \ref{Ex:Con} (this illustration is not
intended to represent any plausible natural language theory, but does use
the ``peak/margin'' terminology sometimes employed in syllable theories).
The set of inputs is \{C,V\}$^{+}$. The candidate descriptions of an input
consist of a sequence of pieces, each of which has a peak (p) surrounded by
one or more pairs of margin positions (m). These structures exhibit
prototypical context-free behavior, in that margin positions to the left of
a peak are balanced with margin positions to the right. `e' is the empty
string, and `S' the start symbol.

\begin{example}
\label{Ex:PSG}The Position Structure Grammar
\end{example}

\begin{tabular}{ccl}
S & $\Rightarrow $ & F $|$ e \\ 
F & $\Rightarrow $ & Y $|$ YF \\ 
Y & $\Rightarrow $ & P $|$ MFM \\ 
M & $\Rightarrow $ & m \\ 
P & $\Rightarrow $ & p
\end{tabular}

\begin{example}
\label{Ex:Con}The Constraints
\end{example}

\begin{tabular}{ll}
{-(}m/V) & Do not parse V into a margin position \\ 
{-(}p/C) & Do not parse C into a peak position \\ 
{\sc Parse} & Input segments must be parsed \\ 
{\sc Fill}$^m$ & A margin position must be filled \\ 
{\sc Fill}$^p$ & A peak position must be filled
\end{tabular}

The first two constraints are structural, and mandate that V not be parsed
into a margin position, and that C not be parsed into a peak position. The
other three constraints are faithfulness constraints. The two structural
constraints are satisfied by descriptions with each V in a peak position
surrounded by matched C's in margin positions: CCVCC, V, CVCCCVCC, etc. If
the input string permits such an analysis, it will be given this completely
faithful description, with no resulting constraint violations (ensuring that
it will be optimal with respect to any ranking).

Consider the constraint hierarchy in Example \ref{Ex:Hierarchy}.

\begin{example}
\label{Ex:Hierarchy}A Constraint Hierarchy
\end{example}

\begin{center}
\{-(m/V), -(p/C), {\sc Parse}\} $\gg $ \{{\sc Fill}$^p$\} $\gg $ \{{\sc Fill}%
$^m$\}
\end{center}

This ranking ensures that in optimal descriptions, a V will only be parsed
as a peak, while a C will only be parsed as a margin. Further, all input
segments will be parsed, and unfilled positions will be included only as
necessary to produce a sequence of balanced structures. For example, the
input /VC/ receives the description\footnote{%
In this paper, tree structures will be denoted with parentheses: a parent
node X with child nodes Y and Z is denoted X(Y,Z).} shown in Example \ref
{Ex:OptDesc}.

\begin{example}
\label{Ex:OptDesc}The Optimal Description for /VC/
\end{example}

\begin{center}
S(F(Y(M(${\cal C}$),P(V),M(C))))
\end{center}

The surface string for this description is ${\cal C}$VC: the first ${\cal C}$
was ``epenthesized'' to balance with the one following the peak V. This
candidate is optimal because it only violates {\sc Fill}$^m$, the
lowest-ranked constraint.

Tesar identifies locality as a sufficient condition on the universal
constraints for the success of his approach. For formally regular position
structure grammars, he defines a local constraint as one which can be
evaluated strictly on the basis of two consecutive positions (and any input
segments filling those positions) in the linear position structure. That
idea can be extended to the context-free case as follows. A local constraint
is one which can be evaluated strictly on the basis of the information
contained within a local region. A local region of a description is either
of the following:

\begin{itemize}
\item  a non-terminal and the child non-terminals that it immediately
dominates;
\end{itemize}

\begin{itemize}
\item  a non-terminal which dominates a terminal symbol (position), along
with the terminal and the input segment (if present) filling the terminal
position.
\end{itemize}

It is important to keep clear the role of the position structure grammar. It
does not define the set of grammatical structures, it defines the space of
candidate structures. Thus, the computation of descriptions addressed in
this paper should be distinguished from robust, or error-correcting, parsing 
\cite[ for example]{Anderson+Backhouse:parse-error}. There, the input string
is mapped to the grammatical structure that is `closest'; if the input
completely matches a structure generated by the grammar, that structure is
automatically selected. In the OT case presented here, the full grammar is
the entire OT system, of which the position structure grammar is only a
part. Error-correcting parsing uses optimization only with respect to the
faithfulness of pre-defined grammatical structures to the input. OT uses
optimization to define grammaticality.

\section{The Dynamic Programming Table}

The Dynamic Programming (DP) Table is here a three-dimensional,
pyramid-shaped data structure. It resembles the tables used for context-free
chart parsing \cite{Kay:ChartParsing} and maximum likelihood computation for
stochastic context-free grammars \cite{Lari+Young:SCFG} \cite
{Charniak:book93}. Each cell of the table contains a partial description (a
part of a structural description), and the Harmony of that partial
description. A partial description is much like an edge in chart parsing,
covering a contiguous substring of the input. A cell is identified by three
indices, and denoted with square brackets (e.g., [X,a,c]). The first index
identifying the cell (X) indicates the cell category of the cell. The other
two indices (a and c) indicate the contiguous substring of the input string
covered by the partial description contained in the cell (input segments $%
i_a $ through $i_c$).

In chart parsing, the set of cell categories is precisely the set of
non-terminals in the grammar, and thus a cell contains a subtree with a root
non-terminal corresponding to the cell category, and with leaves that
constitute precisely the input substring covered by the cell. In the
algorithm presented here, the set of cell categories are the non-terminals
of the position structure grammar, along with a category for each
left-aligned substring of the right hand side of each position grammar rule.
Example \ref{Ex:CellCategories} gives the set of cell categories for the
position structure grammar in Example \ref{Ex:PSG}.

\begin{example}
\label{Ex:CellCategories}The Set of Cell Categories
\end{example}

\begin{center}
S, F, Y, M, P, MF
\end{center}

The last category in Example \ref{Ex:CellCategories}, MF, comes from the
rule Y $\Rightarrow $ MFM of Example \ref{Ex:PSG}, which has more than two
non-terminals on the right hand side. Each such category corresponds to an
incomplete edge in normal chart parsing; having a table cell for each such
category eliminates the need for a separate data structure containing edges.
The cell [MF,a,c] may contain an ordered pair of subtrees, the first with
root M covering input [a,b], and the second with root F covering input
[b+1,c].

The DP Table is perhaps best envisioned as a set of layers, one for each
category. A layer is a set of all cells in the table indexed by a particular
cell category.

\begin{example}
A Layer of the Dynamic Programming Table for Category M (input $i_1$-$i_3$)
\end{example}

\begin{tabular}{ccc}
\cline{1-1}
\multicolumn{1}{|c}{[M,1,3]} & \multicolumn{1}{|c}{} &  \\ \cline{1-2}
\multicolumn{1}{|c}{[M,1,2]} & \multicolumn{1}{|c}{[M,2,3]} & 
\multicolumn{1}{|c}{} \\ \hline
\multicolumn{1}{|c}{[M,1,1]} & \multicolumn{1}{|c}{[M,2,2]} & 
\multicolumn{1}{|c|}{[M,3,3]} \\ \hline
$i_1$ & $i_2$ & $i_3$%
\end{tabular}

For each substring length, there is a collection of rows, one for each
category, which will collectively be referred to as a {\em level}. The first
level contains the cells which only cover one input segment; the number of
cells in this level will be the number of input segments multiplied by the
number of cell categories. Level two contains cells which cover input
substrings of length two, and so on. The top level contains one cell for
each category. One other useful partition of the DP table is into blocks. A
block is a set of all cells covering a particular input subsequence. A block
has one cell for each cell category.

A cell of the DP Table is filled by comparing the results of several
operations, each of which try to fill a cell. The operation producing the
partial description with the highest Harmony actually fills the cell. The
operations themselves are discussed in Section \ref{Sec:OpSet}.

The algorithm presented in Section 6 fills the table cells level by level:
first, all the cells covering only one input segment are filled, then the
cells covering two consecutive segments are filled, and so forth. When the
table has been completely filled, cell [S,1,J] will contain the optimal
description of the input, and its Harmony. The table may also be filled in a
more left-to-right manner, bottom-up, in the spirit of CKY. First, the cells
covering only segment $i_1$, and then $i_2$, are filled. Then, the cells
covering the first two segments are filled, using the entries in the cells
covering each of $i_1$ and $i_2$. The cells of the next diagonal are then
filled.

\section{The Operations Set\label{Sec:OpSet}}

The Operations Set contains the operations used to fill DP Table cells. The
algorithm proceeds by considering all of the operations that could be used
to fill a cell, and selecting the one generating the partial description
with the highest Harmony to actually fill the cell. There are three main
types of operations, corresponding to underparsing, parsing, and overparsing
actions. These actions are analogous to the three primitive actions of
sequence comparison \cite{Sankoff+Kruskal:DP}: deletion, correspondence, and
insertion.

The discussion that follows makes the assumption that the right hand side of
every production is either a string of non-terminals or a single terminal.
Each parsing operation generates a new element of structure, and so is
associated with a position structure grammar production. The first type of
parsing operation involves productions which generate a single terminal
(e.g., P$\Rightarrow $p). Because we are assuming that an input segment may
only be parsed into at most one position, and that a position may have at
most one input segment parsed into it, this parsing operation may only fill
a cell which covers exactly one input segment. in our example, cell [P,1,1]
could be filled by an operation parsing $i_1$ into a p position, giving the
partial description P(p filled with $i_1$).

The other kinds of parsing operations are matched to position grammar
productions in which a parent non-terminal generates child non-terminals.
One of these kinds of operations fills the cell for a category by combining
cell entries for two factor categories, in order, so that the substrings
covered by each of them combine (concatenatively, with no overlap) to form
the input substring covered by the cell being filled. For rule Y $%
\Rightarrow $ MFM, there will be an operation of this type combining entries
in [M,a,b] and [F,b+1,c], creating the concatenated structure\footnote{%
This partial description is not a single tree, but an ordered pair of trees.
In general, such concatenated structures will be ordered lists of trees.}
[M,a,b]+[F,b+1,c], to fill [MF,a,c]. The final type of parsing operation
fills a cell for a category which is a single non-terminal on the left hand
side of a production, by combining two entries which jointly form the entire
right hand side of the production. This operation would combining entries in
[MF,a,c] and [M,c+1,d], creating the structure Y([MF,a,c],[M,c+1,d]), to
fill [Y,a,d]. Each of these operations involves filling a cell for a target
category by using the entries in the cells for two factor categories.

The resulting Harmony of the partial description created by a parsing
operation will be the combination of the marks assessed each of the partial
descriptions for the factor categories, plus any additional marks incurred
as a result of the structure added by the production itself. This is true
because the constraints must be local: any new constraint violations are
determinable on the basis of the cell category of the factor partial
descriptions, and not any other internal details of those partial
descriptions.

All possible ways in which the factor categories, taken in order, may
combine to cover the substring, must be considered. Because the factor
categories must be contiguous and in order, this amounts to considering each
of the ways in which the substring can be split into two pieces. This is
reflected in the parsing operation descriptions given in Section \ref
{Subsec:Ops}.

Underparsing operations are not matched with position grammar productions. A
DP Table cell which covers only one input segment may be filled by an
underparsing operation which marks the input segment as underparsed. In
general, any partial description covering any substring of the input may be
extended to cover an adjacent input segment by adding that additional
segment marked as underparsed. Thus, a cell covering a given substring of
length greater than one may be filled in two mirror-image ways via
underparsing: by taking a partial description which covers all but the
leftmost input segment and adding that segment as underparsed, and by taking
a partial description which covers all but the rightmost input segment and
adding that segment as underparsed.

Overparsing operations are discussed in Section \ref{Sec:OverParsing}.

\section{The Overparsing Operations\label{Sec:OverParsing}}

Overparsing operations consume no input; they only add new unfilled
structure. Thus, a block of cells (the set of cells each covering the same
input substring) is interdependent with respect to overparsing operations,
meaning that an overparsing operation trying to fill one cell in the block
is adding structure to a partial description from a different cell in the
same block. The first consequence of this is that the overparsing operations
must be considered after the underparsing and parsing operations for that
block. Otherwise, the cells would be empty, and the overparsing operations
would have nothing to add on to.

The second consequence is that overparsing operations may need to be
considered more than once, because the result of one overparsing operation
(if it fills a cell) could be the source for another overparsing operation.
Thus, more than one pass through the overparsing operations for a block may
be necessary. In the description of the algorithm given in Section \ref
{Subsec:MainAlg}, each Repeat-Until loop considers the overparsing
operations for a block of cells. The number of loop iterations is the number
of passes through the overparsing operations for the block. The loop
iterations stop when none of the overparsing operations is able to fill a
cell (each proposed partial description is less harmonic than the partial
description already in the cell).

In principle, an unbounded number of overparsing operations could apply, and
in fact descriptions with arbitrary numbers of unfilled positions are
contained in the output space of {\it Gen} (as formally defined). The
algorithm does not have to explicitly consider arbitrary amounts of
overparsing, however. A necessary property of the faithfulness constraints,
given constraint locality, is that a partial description cannot have
overparsed structures repeatedly added to it until the resulting partial
description falls into the same cell category as the original prior to
overparsing, and be more Harmonic. Such a sequence of overparsing operations
can be considered a {\em overparsing cycle}. Thus, the faithfulness
constraints must ban overparsing cycles. This is not solely a computational
requirement, but is necessary for the grammar to be well-defined:
overparsing cycles must be harmonically suboptimal, otherwise arbitrary
amounts of overparsing will be permitted in optimal descriptions. In
particular, the constraints should prevent overparsing from adding an entire
overparsed non-terminal more than once to the same partial description while
passing through the overparsing operations. In Example \ref{Ex:Con}, the
constraints {\sc Fill}$^m$ and {\sc Fill}$^p$ effectively ban overparsing
cycles: no matter where these constraints are ranked, a description
containing an overparsing cycle will be less harmonic (due to additional 
{\sc Fill} violations) than the same description with the cycle removed.

Given that the universal constraints meet this criterion, the overparsing
operations may be repeatedly considered for a given level until none of them
increase the Harmony of the entries in any of the cells. Because each
overparsing operation maps a partial description in one cell category to one
for another cell category, a partial description cannot undergo more
consecutive overparsing operations than there are cell categories without
repeating at least one cell category, thereby creating a cycle. Thus, the
number of cell categories places a constant bound on the number of passes
made through the overparsing operations for a block.

A single non-terminal may dominate an entire subtree in which none of the
syllable positions at the leaves of the tree are filled. Thus, the optimal
``unfilled structure'' for each non-terminal, and in fact each cell
category, must be determined, for use by the overparsing operations. The
optimal overparsing structure for category X is denoted with [X,0], and such
an entity is referred to as a {\em base overparsing structure}. A set of
such structures must be computed, one for each category, before filling
input-dependent DP table cells. Because these values are not dependent upon
the input, base overparsing structures may be computed and stored in
advance. Computing them is just like computing other cell entries, except
that only overparsing operations are considered. First, consider (once) the
overparsing operations for each non-terminal X which has a production rule
permitting it to dominate a terminal x: each tries to set [X,0] to contain
the corresponding partial description with the terminal x left unfilled.
Next consider the other overparsing operations for each cell, choosing the
most Harmonic of those operations' partial descriptions and the prior value
of [X,0].

\section{The Dynamic Programming Algorithm\label{Sec:DPAlg}}

\subsection{Notation}

maxH\{\} returns the argument with maximum Harmony

\noindent $\left\langle i_a\right\rangle $ denotes input segment $i_a$
underparsed

\noindent X$^t$ is a non-terminal

\noindent x$^t$ is a terminal

\noindent + denotes concatenation

\subsection{The Operations\label{Subsec:Ops}}

Underparsing Operations for [X$^t$,a,a]:

create $\left\langle i_a\right\rangle $+[X$^t$,0]\\

\noindent Underparsing Operations for [X$^t$,a,c]:

create $\left\langle i_a\right\rangle $+[X$^t$,a+1,c]

create [X$^t$,a,c-1]+$\left\langle i_a\right\rangle $\\

\noindent Parsing operations for [X$^t$,a,a]:

for each production X$^t\Rightarrow $ x$^k$

\hspace{.125in}create X$^t$(x$^k$ filled with i$_a$)\\

\noindent Parsing operations for [X$^t$,a,c],

\noindent where c$>$a and all X are cell categories:

for each production X$^t\Rightarrow $ X$^k$X$^m$

\hspace{.125in}for b = a+1 to c-1

\hspace{.125in}\hspace{.125in}create X$^t$([X$^k$,a,b],[X$^m$,b+1,c])

for each production X$^u\Rightarrow $ X$^k$X$^m$X$^n$...

where X$^t$ = X$^k$X$^m$:

\hspace{.125in}for b=a+1 to c-1

\hspace{.125in}\hspace{.125in}create [X$^k$,a,b]+[X$^m$,b+1,c]\\

\noindent Overparsing operations for [X$^t$,0]:

for each production X$^t\Rightarrow $ x$^k$

\hspace{.125in}create X$^t$(x$^k$ unfilled)

for each production X$^t\Rightarrow $ X$^k$X$^m$

\hspace{.125in}create X$^t$([X$^k$,0],[X$^m$,0])

for each production X$^u\Rightarrow $ X$^k$X$^m$X$^n$...

where X$^t$ = X$^k$X$^m$:

\hspace{.125in}create [X$^k$,0]+[X$^m$,0]\\

\noindent Overparsing operations for [X$^t$,a,a]:

same as for [X$^t$,a,c]\\

\noindent Overparsing operations for [X$^t$,a,c]:

for each production X$^t\Rightarrow $ X$^k$

\hspace{.125in}create X$^t$([X$^k$,a,c])

for each production X$^t\Rightarrow $ X$^k$X$^m$

\hspace{.125in}create X$^t$([X$^k$,0],[X$^m$,a,c])

\hspace{.125in}create X$^t$([X$^k$,a,c],[X$^m$,0])

for each production X$^u\Rightarrow $ X$^k$X$^m$X$^n$...

where X$^t$ = X$^k$X$^m$:

\hspace{.125in}create [X$^k$,a,c]+[X$^m$,0]

\hspace{.125in}create [X$^k$,0]+[X$^m$,a,c]

\subsection{The Main Algorithm\label{Subsec:MainAlg}}

\noindent /* create the base overparsing structures */\\

\noindent Repeat

For each X$^t$, Set [X$^t$,0] to

maxH\{[X$^t$,0], overparsing ops for [X$^t$,0]\}

\noindent Until no [X$^t$,0] has changed during a pass\\

\noindent /* fill the cells covering only a single segment */\\

\noindent For a = 1 to J

For each X$^t$, Set [X$^t$,a,a] to

maxH\{underparsing ops for [X$^t$,a,a]\}

For each X$^t$, Set [X$^t$,a,a] to

maxH\{[X$^t$,a,a], parsing ops for [X$^t$,a,a]\}

Repeat

\hspace{.125in}For each X$^t$, Set [X$^t$,a,a] to

\hspace{.125in}maxH\{[X$^t$,a,a], overparsing ops for [X$^t$,a,a]\}

Until no [X$^t$,a,a] has changed during a pass\\

\noindent /* fill the rest of the cells */\\

\noindent For d=1 to (J-1)

For a=1 to (J-d)

\hspace{.125in}For each X$^t$, Set [X$^t$,a,a+d] to

\hspace{.125in}maxH\{underparsing ops for [X$^t$,a,a+d]\}

\hspace{.125in}For each X$^t$, Set [X$^t$,a,a+d]

\hspace{.125in}maxH\{[X$^t$,a,a+d], parsing ops for [X$^t$,a,a+d]\}

\hspace{.125in}Repeat

\hspace{.125in}\hspace{.125in}For each X$^t$,

\hspace{.125in}\hspace{.125in}\hspace{.125in}Set [X$^t$,a,a+d] to

\hspace{.125in}\hspace{.125in}\hspace{.125in}maxH\{[X$^t$,a,a+d],

\hspace{.125in}\hspace{.125in}\hspace{.125in}\hspace{.125in}overparsing ops
for [X$^t$,a,a+d]\}

\hspace{.125in}Until no [X$^t$,a,a+d] has changed during a pass

\noindent Return [S,1,J] as the optimal description

\subsection{Complexity}

Each block of cells for an input subsequence is processed in time linear in
the length of the subsequence. This is a consequence of the fact that in
general parsing operations filling such a cell must consider all ways of
dividing the input subsequence into two pieces. The number of overparsing
passes through the block is bounded from above by the number of cell
categories, due to the fact that overparsing cycles are suboptimal. Thus,
the number of passes is bounded by a constant, for any fixed position
structure grammar. The number of such blocks is the number of distinct,
contiguous input subsequences (equivalently, the number of cells in a
layer), which is on the order of the square of the length of the input. If N
is the length of the input, the algorithm has computational complexity O(N$%
^3 $).

\section{Discussion}

\subsection{Locality}

That locality helps processing should be no great surprise to
computationalists; the computational significance of locality is widely
appreciated. Further, locality is often considered a desirable property of
principles in linguistics, independent of computational concerns.
Nevertheless, locality is a sufficient but not necessary restriction for the
applicability of this algorithm. The locality restriction is really a
special case of a more general sufficient condition. The general condition
is a kind of ``Markov'' property. This property requires that, for any
substring of the input for which partial descriptions are constructed, the
set of possible partial descriptions for that substring may be partitioned
into a finite set of classes, such that the consequences in terms of
constraint violations for the addition of structure to a partial description
may be determined entirely by the identity of the class to which that
partial description belongs. The special case of strict locality is easy to
understand with respect to context-free structures, because it states that
the only information needed about a subtree to relate it to the rest of the
tree is the identity of the root non-terminal, so that the (necessarily
finite) set of non-terminals provides the relevant set of classes.

\subsection{Underparsing and Derivational Redundancy}

The treatment of the underparsing operations given above creates the
opportunity for the same partial description to be arrived at through
several different paths. For example, suppose the input is $%
i_a...i_bi_ci_d...i_e$ , and there is a constituent in [X,a,b] and a
constituent [Y,d,e]. Further suppose the input segment $i_c$ is to be marked
underparsed, so that the final description [S,a,e] contains [X,a,b] $%
\left\langle i_c\right\rangle $ [Y,d,e]. That description could be arrived
at either by combining [X,a,b] and $\left\langle i_c\right\rangle $ to fill
[X,a,c], and then combine [X,a,c] and [Y,d,e], or it could be arrived at by
combining $\left\langle i_c\right\rangle $ and [Y,d,e] to fill [Y,c,e], and
then combine [X,a,b] and [Y,c,e]. The potential confusion stems from the
fact that an underparsed segment is part of the description, but is not a
proper constituent of the tree.

This problem can be avoided in several ways. An obvious one is to only
permit underparsings to be added to partial descriptions on the right side.
One exception would then have to be made to permit input segments prior to
any parsed input segments to be underparsed (i.e., if the first input
segment is underparsed, it has to be attached to the left side of some
constituent because it is to the left of everything in the description).

\section{Conclusions}

The results presented here demonstrate that the basic cubic time complexity
results for processing context-free structures are preserved when Optimality
Theory grammars are used. If {\it Gen} can be specified as matching input
segments to structures generated by a context-free position structure
grammar, and the constraints are local with respect to those structures,
then the algorithm presented here may be applied directly to compute optimal
descriptions.

\section{Acknowledgments}

I would like to thank Paul Smolensky for his valuable contributions and
support. I would also like to thank David Haussler, Clayton Lewis, Mark
Liberman, Jim Martin, and Alan Prince for useful discussions, and three
anonymous reviewers for helpful comments. This work was supported in part by
an NSF Graduate Fellowship to the author, and NSF grant IRI-9213894 to Paul
Smolensky and Geraldine Legendre.


\begin{thebibliography}{fullname}
\bibitem[\protect\citename{Anderson and Backhouse}1981]{Anderson+Backhouse:parse-error}  %
S. O. Anderson and R. C. Backhouse. 1981. Locally least-cost error recovery
in Earley's algorithm. {\em ACM Transactions on Programming Languages and
Systems} 3: 318-347.

\bibitem[\protect\citename{Charniak}1993]{Charniak:book93}  Eugene Charniak.
1993. {\em Statistical language learning}. Cambridge, MA: MIT Press.

\bibitem[\protect\citename{Kay}1980]{Kay:ChartParsing}  Martin Kay. 1980.
Algorithmic schemata and data structures in syntactic processing. CSL-80-12,
October 1980.

\bibitem[\protect\citename{Lari and Young}1990]{Lari+Young:SCFG}  K. Lari
and S. J. Young. 1990. The estimation of stochastic context-free grammars
using the inside-outside algorithm. {\em Computer Speech and Language} 4:
35-36.

\bibitem[\protect\citename{Lewis and Papadimitriou}1981]{L+P:TheoryComp}  %
Harry R. Lewis and Christos H. Papadimitriou. 1981. {\em Elements of the
theory of computation}. Englewood Cliffs, New Jersey: Prentice-Hall, Inc.

\bibitem[\protect\citename{Prince and Smolensky}1993]{Prince+Smolensky:OT}  %
Alan Prince and Paul Smolensky. 1993. {\em Optimality Theory: Constraint
interaction in generative grammar}. Technical Report CU-CS-696-93,
Department of Computer Science, University of Colorado at Boulder, and
Technical Report TR-2, Rutgers Center for Cognitive Science, Rutgers
University, New Brunswick, NJ. March. To appear in the Linguistic Inquiry
Monograph Series, Cambridge, MA: MIT Press.

\bibitem[\protect\citename{Sankoff and Kruskal}1983]{Sankoff+Kruskal:DP}  %
David Sankoff and Joseph Kruskal. 1983. {\em Time warps, string edits, and
macromolecules: The theory and practice of sequence comparison}. Reading,
MA: Addison-Wesley.

\bibitem[\protect\citename{Tesar}1994]{Tesar:ParsingTR94}  Bruce Tesar.
1994. Parsing in Optimality Theory: A dynamic programming approach.
Technical Report CU-CS-714-94, April 1994. Department of Computer Science,
University of Colorado, Boulder.

\bibitem[\protect\citename{Tesar}1995a]{Tesar:ParsingTR95}  Bruce Tesar.
1995a. Computing optimal forms in Optimality Theory: Basic syllabification.
Technical Report CU-CS-763-95, February 1995. Department of Computer
Science, University of Colorado, Boulder.

\bibitem[\protect\citename{Tesar}1995b]{Tesar:Dissertation}  Bruce Tesar.
1995b. Computational Optimality Theory. Unpublished Ph.D. Dissertation.
Department of Computer Science, University of Colorado, Boulder. June 1995.

\bibitem[\protect\citename{Viterbi}1967]{Viterbi:VAlg}  A.J. Viterbi. 1967.
Error bounds for convolution codes and an asymptotically optimal decoding
algorithm. {\em IEEE Trans. on Information Theory} 13:260-269.
\end{thebibliography}
\end{document}